\def\year{2022}\relax
\documentclass[letterpaper]{article} % DO NOT CHANGE THIS
\usepackage{aaai22}  % DO NOT CHANGE THIS
\usepackage{times}  % DO NOT CHANGE THIS
\usepackage{helvet}  % DO NOT CHANGE THIS
\usepackage{courier}  % DO NOT CHANGE THIS
\usepackage[hyphens]{url}  % DO NOT CHANGE THIS
\usepackage{graphicx} % DO NOT CHANGE THIS
\usepackage{natbib}  % DO NOT CHANGE THIS AND DO NOT ADD ANY OPTIONS TO IT
\usepackage{caption} % DO NOT CHANGE THIS AND DO NOT ADD ANY OPTIONS TO IT

\usepackage[pagewise]{lineno}
\usepackage{lipsum}
\usepackage{multirow}
\usepackage{amssymb}
\usepackage[marginal]{footmisc}
\usepackage{enumitem}
\usepackage{amsthm}
\usepackage{amsmath}

\DeclareCaptionStyle{ruled}{labelfont=normalfont,labelsep=colon,strut=off} % DO NOT CHANGE THIS
\usepackage{algorithm}
\usepackage{algorithmic}

\usepackage{newfloat}
\usepackage{listings}
\floatstyle{ruled}
\newfloat{listing}{tb}{lst}{}
\floatname{listing}{Listing}

\title{Unsupervised clothing change adaptive person ReID}
\author{
    Ziyue Zhang, Shuai Jiang, Congzhentao Huang, Richard YiDa Xu
   
}
\affiliations{
    University of Technology Sydney 
}

\usepackage{bibentry}
\begin{document}

\maketitle

\begin{abstract}
Clothing changes and lack of data labels are both crucial challenges in person ReID. For the former challenge, people may occur multiple times at different locations wearing different clothing. However, most of the current person ReID research works focus on the benchmarks in which a person's clothing is kept the same all the time. For the last challenge, some researchers try to make model learn information from a labeled dataset as a source to an unlabeled dataset. Whereas purely unsupervised training is less used. In this paper, we aim to solve both problems at the same time. We design a novel unsupervised model, Sync-Person-Cloud ReID, to solve the unsupervised clothing change person ReID problem. We developer a purely unsupervised clothing change person ReID pipeline with person sync augmentation operation and same person feature restriction. The person sync augmentation is to supply additional same person resources. These same person's resources can be used as part supervised input by same person feature restriction. The extensive experiments on clothing change ReID datasets show the out-performance of our methods.
\end{abstract}

\section{Introduction}

Person re-identification (ReID) \cite{ye2021deep} is designed to match specific pedestrians in images or video sequences.
The main challenge of ReID is that the ReID features variations of the same person in different situations are usually significant because of viewpoint or situation differences, which makes it challenging to identify the same person. Meanwhile, the lack of ReID features variations of different people in the same situation or wearing the same clothing also influence ReID performance. 
Current works in person ReID are mostly to learn discriminative features of person identity by a specifically designed backbone model \cite{chang2018multi,wang2019spatial}. There are also works focusing on problems of occlusions \cite{hou2019vrstc}, different poses \cite{qian2018pose}, illumination changes \cite{9190796} and resolution changes \cite{li2019recover}. 

\begin{figure}[!h]
    \centering
    \includegraphics[width = 0.45\textwidth]{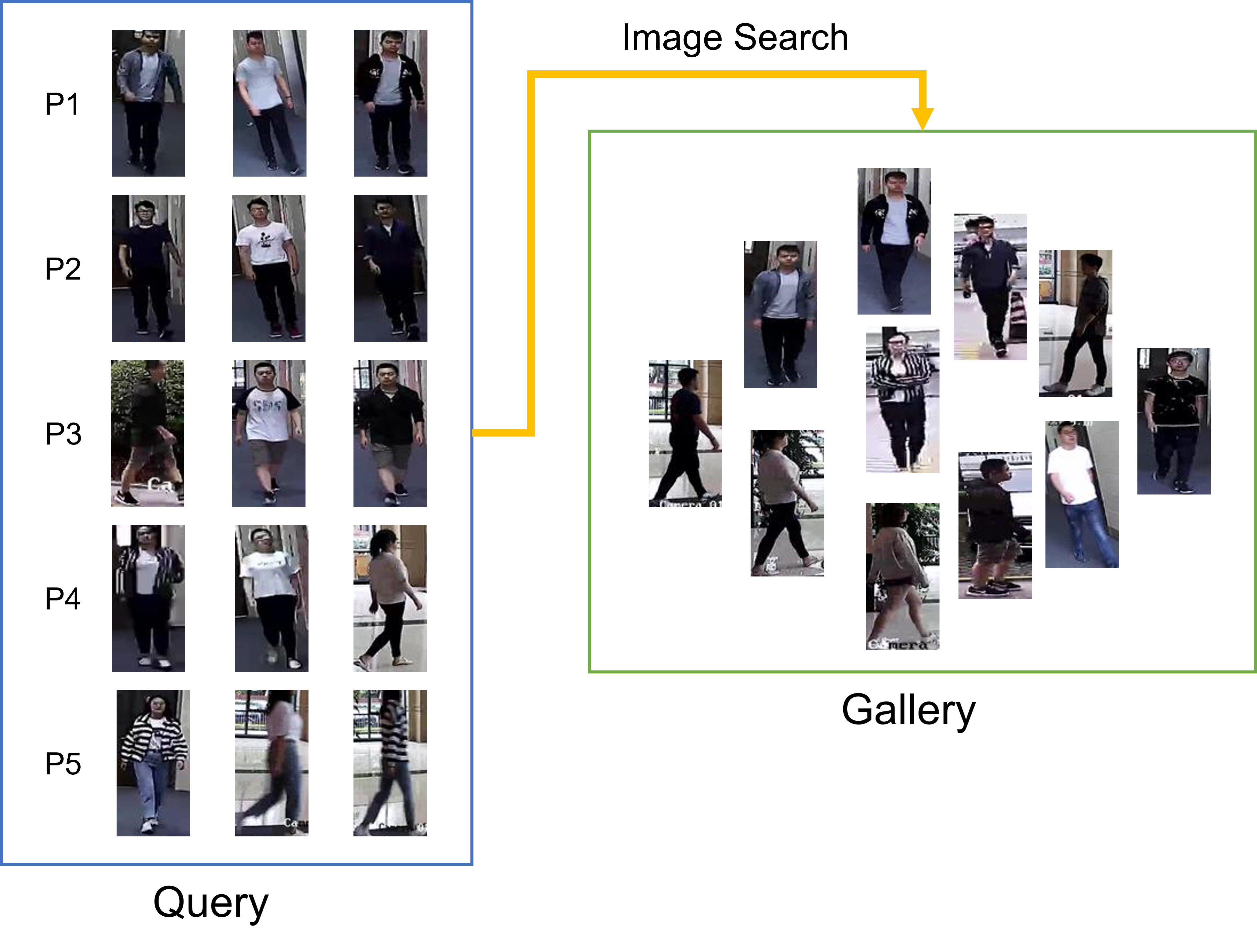}
    \caption{Example of clothing change person ReID dataset Real28 \cite{wan2020person}. The query images include five persons' images. The images of each row in the query are from the same person with different clothes. The gallery images consist of different person images with different clothes.}
    \label{fig:ClothExample}
\end{figure}

The clothing change plays a crucial character in person ReID. However, the researchers assume the same person wears the same clothes, which means the above works cannot handle the clothing change variations in person ReID. In daily life, people usually change their clothes, which means that the same person captured by the camera at different times may wear different clothes. As shown in the figure \ref{fig:ClothExample}, clothing change ReID datasets consist of a person with different clothing and researchers try to find the same person under such settings, which is more in line with the real-life scene. The existing conventional methods tend to fail in such a scenario because of the unreliability of clothing texture information and the lack of the same person's ID information with different clothing.

Another key problem is that it is challenging to get person ReID labels.  Because it is expensive to record person images across multiple cameras, many previous works try to address person ReID problem by unsupervised or semi-supervised learning \cite{lin2019bottom, yu2019unsupervised, wang2018transferable}. 
% There are two kinds of unsupervised ReID methods. One is purely unsupervised learning and the other is unsupervised domain adaptation. The purely unsupervised learning generally exploits pseudo labels from the completely unlabeled data by clustering. 
Most of these methods use unsupervised domain adaptation. They first pretrain a model on the source labeled dataset and then fine-tune it on the target unlabeled dataset. However, these methods still need at least one label dataset and suffer from the variation between the source and target domain. Hence, we choose to use purely unsupervised learning for our method.

In this paper, we propose a pure unsupervised model called Sync-Person-Cloud ReID to solve the unsupervised clothing change person ReID problem. The whole pipeline follows the below procedure. The person image features of the training data are extracted by the CNN backbone. Then we use a clustering algorithm to cluster image features and produce pseudo labels. At last, the backbone is trained with a contrastive loss such as InfoNCE loss using the storage person image features. 
This pipeline can initially obtain pedestrian features by purely unsupervised learning. However, it is still not able to separate the noise information caused by different clothes and the unsupervised features are not robust enough. Because people always change clothes on daily life, it is not possible to create a simple disentanglement classifier. 
So to better bridge the gap between the same person features with different clothing and improve the unsupervised performance, we have added several innovations summarised as follows: 

\begin{enumerate}
\item
We use a clothing change person augmentation module to generate synthetic person images with different clothes. A person image is sent to the augmentation module to get multiple synthetic person images with person parsing and clothing template. By using this scheme, we can get semi-supervised information from synthetic images which can be seemed as the same person.

\item 
To constrain the synthetic same person features, we adopt a self cluster loss to reduce the distance between these features. This will let the later cluster algorithm perform better by constraining the same person images with different clothes as the same pseudo label. The augmentation images can use the semi-supervised information to make the ReID backbone more robust.

\item 
Both the average and batch hard sampling are used on the cluster image features. It takes much fewer GPU resources than using all image features. With the average sampling, the model can be trained by the whole dataset feature to get high performance on a whole large dataset. With batch hard sampling, the model can be trained faster and more accurately. 

% \item
% Extensive experiments on clothing change datasets shows that our method improves the re-ID accuracy in the clothing change unsupervised evaluation.
\end{enumerate}

The rest of the article is organized as follows. Section \textbf{Related work} illustrates related works on conventional person Re-ID, unsupervised person ReID and clothing change person ReID. Section \textbf{Method} describes our pipeline structure for getting unsupervised clothing change person ReID features in detail. Our experiment results and details are shown in section \textbf{Experiments}. We finally conclude our work in section \textbf{Conclusion}. 

\section{Related work}

\subsection{Conventional Person ReID}

Most researchers focused on traditional person ReID. In the early research, one primary method of person ReID is metric learning, which is to formalize the problem as supervised metric learning where a projection matrix is sought out \cite{yang2017person, yi2014deep, ding2015deep}. 
Benefiting from the advances of convolutional neural network (CNN) architectures, another primary method is to learn appropriate features associated with the same ID using features distance information \cite{hermans2017defense} on a backbone module \cite{chang2018multi,si2018dual}, such as Resnet50 \cite{he2016deep}. 
Our work can seem like the second one. We also use a modified Resnet50 as our ReID backbone.

After getting the ReID features, the person re-id is to train the backbone with contrastive loss. The contrastive loss is to reduce the distances between image features of the same person and to increase the distances between the image features of different persons. Several methods employed triplet loss \cite{hermans2017defense,chen2017beyond,song2018mask} to constrain the distances between image triplets. Some researchers also use identity classification loss \cite{zheng2017unlabeled,zhong2018camera} to make the ReID problem an image classification problem.  
Although these works focus on person ReID, they may fail in some circumstances, e.g., when person images are with different clothing. They cannot handle clothing change ReID data, which will lead to low performance.

\subsection{Unsupervised Person ReID}
Early unsupervised Re-ID works are mainly to learn invariant components, i.e., dictionary or metric, whose discriminability or scalability is relatively insufficient. For deeply unsupervised methods, there are two main types of methods.

The first is unsupervised domain adaptation (UDA), which transfers the knowledge on a labeled source dataset to the unlabeled target dataset. Due to the powerful supervision in the source dataset, it can learn enough information for person ReID.
For example, \cite{deng2018image} presents a baseline to translate the labeled images from source to target domain in an unsupervised manner with self-similarity and domain-dissimilarity. \cite{lin2018multi} developed a Multi-task Mid-level Feature Alignment network that can be jointly optimised under the person's identity classification and the attribute learning task with a cross-dataset mid-level feature alignment regularisation term. \cite{wang2018transferable} proposed a model named Transferable Joint Attribute-Identity Deep Learning for simultaneously learning an attribute-semantic and identity discriminative feature representation space transferable to any new target domain without the need for collecting new labeled training data from the target domain. 
Although these works can learn a new unlabeled dataset, the large domain shift is still a challenge for the UDA problem. 

\begin{figure*}[!ht]
    \centering
    \includegraphics[width = 1\textwidth]{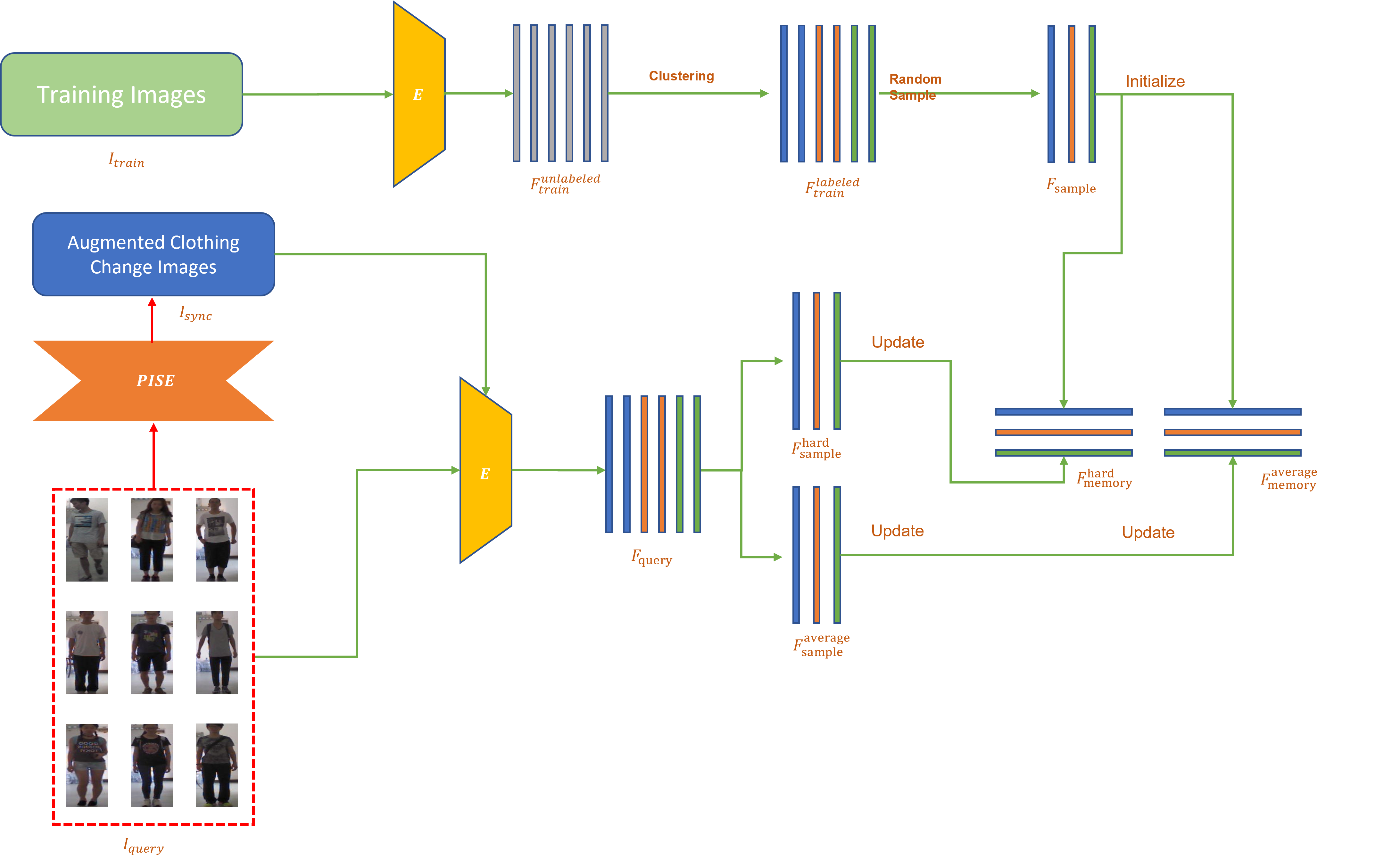}
    \caption{ The whole pipeline of our unsupervised clothing change ReID model (best viewed in color). For conciseness, we do not show the augmentation module $PISE$ details here. }
    \label{fig:pipeline}
\end{figure*}

Another method is end-to-end purely unsupervised learning, which generally generates pseudo labels from the completely unlabeled data using a clustering algorithm. \cite{fan2018unsupervised} proposed an effective baseline named the progressive unsupervised learning (PUL) method to transfer pretrained deep representations to unseen domains. \cite{fu2019self} proposed a self-similarity grouping approach to get the pseudo identities by exploiting the potential similarity of unlabeled samples to build multiple clusters from different views automatically. \cite{lin2019bottom} proposed a bottom-up clustering approach to jointly optimize a convolutional neural network and the relationship among the individual samples. \cite{dai2021cluster} proposed a cluster contrast that stores feature vectors and computes contrast loss in the cluster level to solve the inconsistency problem for cluster feature representation. 

In this paper, we choose to use end-to-end purely unsupervised learning as our initial pipeline.

\subsection{Clothing Change Person ReID}

In a practical surveillance scenario, there are a large number of persons with changing clothes. 
Some researchers use the face and body appearance as the supplement information to address clothing change ReID. \cite{xue2018clothing} proposed a cloth-Clothing Change Aware Network to address the clothing change ReID by separately extracting the face and body context representation. However, the face and body appearance might be unavailable, which may cause the model to fail in some scenarios. 
The infrared or depth images can also be the supplement information to help address clothing change ReID \cite{barbosa2012re,wu2017rgb}. However, the depth information is only applicable in indoor environments, which is not suitable for all scenarios. And the information of the infrared image is much less than RGB images', which makes the ReID less effective.

For purely RGB clothing change person ReID, the methods mostly work on reducing the domain gap between the image features of the same person with different clothes. \cite{hong2021fine} proposed a Fine-grained Shape-Appearance Mutual learning framework that learns fine-grained discriminative body shape knowledge in a shaped stream and transfers it to an appearance stream to complement the cloth-unrelated knowledge in the appearance features. \cite{wan2020person} proposed a new clothing change dataset with both real and synthetic images. \cite{shu2021semantic} proposed a semantic-guided pixel sampling model to automatically learn cloth-irrelevant cues. \cite{yang2019person} proposed a new dataset and present a spatial polar transformation to learn cross-cloth invariant representation.

In this paper, we propose a clothing change augmentation and self identity image constrain to help our unsupervised model solve the clothing change ReID problem

\subsection{Data Augmentation in Person ReID}
Ordinary data augmentations such as random resizing, cropping, random erasing and horizontal flipping are widely used in Re-ID. Besides, using GAN to generates augmented images is also applied in some methods. \cite{zheng2017unlabeled} first attempted to use the GAN for person Re-ID. It improves the supervised feature representation learning with the generated person images. 
\cite{liu2018pose} introduced pose constraints to improve the quality of the generated person images, generating the different pose person images.
\cite{zhong2018camera} add camera style information in the image generation process to address the cross camera adaptation problem. 
Some works also use GAN to generate synthetic person images from one domain to another. \cite{wang2019rgb, zhang2021rgb} used GAN to generate the same content person image from RGB modality to IR modality. \cite{wang2018cascaded} use GAN to transfer low resolution images to high resolution versions.

In our work, we use a pretrained Decoupled GAN proposed by \cite{Zhang_2021_CVPR} to generate different clothes person images. These images are then seemed as same person images by the same person constrain loss.

\section{Method}

\subsection{Overview of Pipeline}

To get the unsupervised clothing change person ReID feature, inspired by \cite{dai2021cluster}, we proposed our unsupervised pipeline for clothing change ReID in a similar way.
We show our whole model pipeline in Figure \ref{fig:pipeline}.

First, we use a modified resnet-50 \cite{he2016deep} as our backbone module $E$ for ReID feature extraction, which is pretrained on ImageNet \cite{deng2009imagenet}. All train dataset images $I_{train}$ will be sent to $E$ to get unlabeled training dataset ReID features $F_{train}^{unlabeled}$.
\begin{equation}
\label{trainInput}
\begin{aligned}
F_{train}^{unlabeled} = E(I_{train})
\end{aligned}
\end{equation}

Then, we use a clustering algorithm DBScan \cite{ester1996density} to generate pseudo labels for each input image feature. DBScan is a density-based clustering method, whose clustering results almost independently depend on the sequence of nodes. It can discover the cluster of any shape and effectively discover noise points, which is more suitable in person ReID problem scenario than other clustering algorithms such as Kmeans \cite{macqueen1967some} and agglomerative clustering \cite{day1984efficient}.
DBScan requires two hyper-parameters. The first is the maximum distance $\varepsilon$, which represents the neighbor radius of the definition density. In other words, $\varepsilon$ is the distance between two samples for one to be considered as in the neighborhood of the other. Another is the cluster neighbor density threshold $M$, which donates the minimum number of samples in a neighborhood for an instance to be considered as a core instance. We set $\varepsilon$ to 0.4 and $M$ to 4 in all our experiments. 
Through DBScan clustering, the cluster ID is assigned to each training image as the pseudo label to get the pseudo-labeled training dataset ReID features.

\begin{equation}
\label{clustering}
\begin{aligned}
F_{train}^{labeled} = DBScan_{\varepsilon}^{M}(F_{train}^{unlabeled}).
\end{aligned}
\end{equation}

Finally, a contrastive loss with the global and average sample is used to compute the loss values between the query instances and the memory dictionary. The details of the computation and other components will be illustrated later.

\subsection{Clothing Change Person Image Augmentation}
\begin{figure}[!hbp]
    \centering
    \includegraphics[width = 0.4\textwidth]{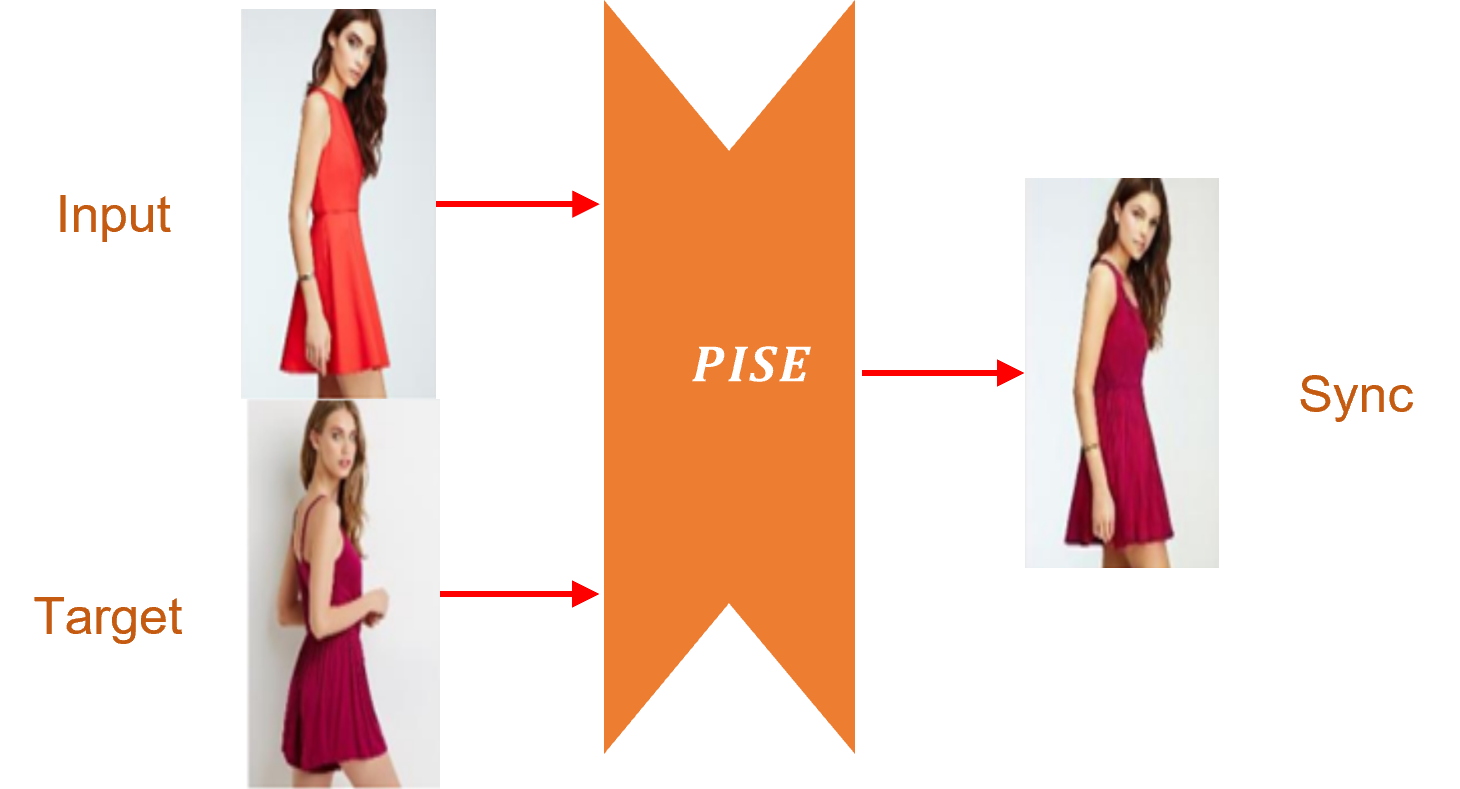}
    \caption{The function of the $PISE$ module. It takes two inputs which are original input query person image and target clothing style image. It will generate a image with the same content of input query person image and same clothing style of input target clothing style image. The synthetic image can be assumed as the same person identity with original input query person image.}
    \label{fig:PISE}
\end{figure}

To get the same person features with different clothes, we introduce a pretrained Person Image Synthesis module $PISE$ \cite{Zhang_2021_CVPR} here. $PISE$ is a two-stage generative model for Person Image Synthesis and Editing, which is able to generate realistic person images with desired poses, textures, or semantic layouts. The effectiveness and function of $PISE$ are shown in Figure \ref{fig:PISE}.

We use $PISE$ to get the synthetic person images $I_{sync}$ from each input query image $I_{query}$. 

\begin{equation}
\label{PISE}
\begin{aligned}
I_{sync} = PISE(I_{query}).
\end{aligned}
\end{equation}

We use $S$ different clothing style template for each input query image to get the corresponding $S$ synthetic images. In our settings, $S$ is set to 4. These synthetic images can be assumed to have the same identity but different clothes. 

\subsection{Cluster Contrast and Update Procedure}

In the very first step of each epoch training, we random sample one single image feature from each clustered pseudo-labeled features $F_{train}^{labeled}$ using uniform sample $U$. 

\begin{equation}
\label{initialize}
\begin{aligned}
F_{sample}(i) = U(F_{train}^{labeled}(i)).
\end{aligned}
\end{equation} 
We donate the $N$ as the cluster number and $F_{sample}(i)$ as a single feature from $i_{th}$ cluster features $F_{train}^{labeled}(i)$. We use these sampled features as the initial memory features for $F_{memory}^{average}$ and $F_{memory}^{hard}$.

During the model training stage, we set the number for query image $I_{query}$ as $P*K$, in which $P$ is person identity number and $K$ is the instance number for each person identity. We send these $K$ person images to the $PISE$ module to get the corresponding $S*K$ synthetic person images $I_{sync}$ with different clothes for each query person. All these images are sent to the ReID backbone to get the query features $F_{query}$.

\begin{equation}
\label{queryfeature}
\begin{aligned}
F_{query} = E(I_{query} \cup I_{sync}).
\end{aligned}
\end{equation}

For updating the feature memory, inspired by \cite{concha2011hmm} which handle the variant dimensions of probability features using normalization, we select the hardest instance and average instance for each person identity with momentum $m$. We maintain two features memory, which is average feature memory $F_{memory}^{average}$ and hardest feature memory $F_{memory}^{hard}$. For a certain cluster with person identity $i$ in hardest feature memory $F_{memory}^{hard}$, its feature vector is updated by:

\begin{equation}
\begin{aligned}
\label{UpdateH}
& f_{query}^{\text {hard }} = \underset{f_{query}}{\arg \max } KL(f_{query} , F_{memory}^{hard}(i)),\\
& f_{query}  \in F_{query}(i), \\
& F_{memory}^{hard}(i) = m \cdot F_{memory}^{hard}(i)+(1-m) \cdot f_{query}^{\text {hard }},
\end{aligned}
\end{equation}
where the batch hard instance $f_{query}^{\text {hard }}$ is the instance with the minimum similarity to the cluster feature. We measure the similarity with KL divergence. $m$ is the momentum updating factor. $F_{query}(i)$ is the instance features set with cluster ID $i$ in the current mini batch.

For a certain cluster with person identity $i$ in average feature memory $F_{memory}^{average}$, its feature vector is updated by:

\begin{equation}
\begin{aligned}
\label{UpdateA}
& f_{\text {query }}^{\text {average }} =\frac{\sum{f_{query}}}{(S+1) \times K}, \\
& f_{query} \in F_{query}(i), \\
& F_{memory}^{average}(i) = m \cdot F_{memory}^{average}(i)+(1-m) \cdot f_{query}^{\text {average}},
\end{aligned}
\end{equation}
where the batch hard instance $f_{query}^{\text {average }}$ is the average instance of all query features $F_{query}(i)$ from identity $i$.

The query image features (including the original input and synthetic image features) are compared to all cluster features (including average memory and hardest memory) with InfoNCE loss \cite{oord2018representation}. The InfoNCE loss can be formulated as follows:

\begin{equation}
\begin{aligned}
\label{InfoNCE}
& L_{q}(F_{query}, F_{memory})=-\log \frac{\exp \left(f_{q} \cdot f_{m}^{+}\right) / \tau}{\sum_{i=0}^{K} \exp \left(f_{q}, f_{m}(i)\right) / \tau}, \\
& f_{q} \in F_{query}, f_{m} \in F_{memory}^{hard} \cup F_{memory}^{average}.
\end{aligned}
\end{equation}

where $f_{m}^{+}$ is the positive memory cluster feature vector to query instance $f_{q}$. $\tau$ is a temperature hyper-parameter \cite{wu2018unsupervised}.
The loss tries to classify $f_{q}$ as $f_{m}^{+}$ through low value when $f_{q}$ is similar to its positive cluster feature and dissimilar to all other cluster features.

\subsection{Self Identity Constrain}

Since the synthetic clothing change images can seem as the same person, we use a self identity constrain loss to reduce the variance between all synthetic person image features generated from the same input person image. We define this loss as follows:

\begin{equation}
\begin{aligned}
\label{SelfI}
& L_{s} =\frac{ \sum_{i=0}^P \sum_{j=0}^K \sum_{a,b=0}^{S+1} KL(f_{query}^a, f_{query}^b)}{P \times K \times (S+1)}, \\
& f_{query} \in F_{query}.
\end{aligned}
\end{equation}

The loss tries to classify all synthetic and corresponding original input person image features as the same person by reducing the KL divergence between them.

\subsection{Whole Train Process}

\begin{algorithm}[h]
\caption{Whole training process for our method}
\label{alg:algorithm}
\begin{algorithmic}[1] 
\REQUIRE  Unlabeled training images $I_{train}$\\
\REQUIRE  Backbone $E$ Pretrained on Imagenet\\
\REQUIRE  Pretrained clothing change person image synthesis module $PISE$\\
\REQUIRE  Hyper parameters $P,K,S,m,\tau,M,\varepsilon$\\
\FOR{$n \in [1, max\_epochs]$}
\STATE Send $I_{train}$ to $E$ to get all features $F_{train}^{unlabeled}$.
\STATE Cluster $F_{train}^{unlabeled}$ to get $F_{train}^{labeled}$ using Eq.\ref{clustering}.
\STATE Initialize the memory $F_{memory}^{average}$ and $F_{memory}^{hard}$ with $F_{sample}$ using Eq.\ref{initialize}.
\FOR{$i \in [1, max\_iter]$}
\STATE Sample $P \times K$ person images $I_{query}$ from $I_{train}$
\STATE Sent $I_{query}$  to $PISE$ to get $I_{sync}$ using Eq.\ref{PISE}.
\STATE Sent $I_{query}$ and $I_{sync}$ to $E$ to get query features $F_{query}$ using Eq.\ref{queryfeature}.
\STATE Compute InfoNCE loss and self identity loss using Eq.\ref{InfoNCE} and Eq.\ref{SelfI} and compute the total loss $L = L_{q} + \alpha L_{s}$.
\STATE Backwards the total loss functions to optimize our encoder $E$.
\STATE Update hardest memory and average memory using Eq.\ref{UpdateH} and Eq.\ref{UpdateA}.

\ENDFOR
\ENDFOR
\end{algorithmic}
\end{algorithm}

We show our whole unsupervised training process in Algorithm \ref{alg:algorithm}. As shown in the pseudo code, the entire training process of each epoch consists of three stages, which are 1) cloud for pseudo label and memory initialization, 2) loss computation for updating backbone and 3) memory update.

\begin{table*}[!ht]
\centering
\caption{Comparison on PRCC and VC-Clothes datasets.The R1 is Rank-1 CMC accuracies (\%). The mAP denotes mean Average Precision score (\%).}
\begin{tabular}{ccccccccc}
\hline
\multirow{3}{*}{Method} & \multicolumn{4}{c}{PRCC}                                                & \multicolumn{4}{c}{VC-Clothes}                                          \\ \cline{2-9} 
                        & \multicolumn{2}{c}{Clothing Change} & \multicolumn{2}{c}{Same Clothing} & \multicolumn{2}{c}{Clothing Change} & \multicolumn{2}{c}{Same Clothing} \\ \cline{2-9} 
                        & R1                & mAP             & R1               & mAP            & R1                & mAP             & R1              & mAP             \\ \hline
LOMO\cite{liao2015person} + KISSME\cite{koestinger2012large}           & 18.55             & -               & 47.40            & -              & -                 & -               & -               & -               \\
LOMO\cite{liao2015person} + XQDA\cite{liao2015person}             & 14.53             & -               & 29.41            & -              & 34.5              & 30.9            & 86.2            & 83.3            \\
PCB\cite{sun2018beyond}                     & 41.8              & 38.7            & 86.88            & -              & 62.0              & 62.2            & 94.7            & 94.3            \\
RGA-SC\cite{zhang2020relation}                  & 42.3              & -               & 98.4             & -              & 71.1              & 67.4            & 95.4            & 94.8            \\
MGN\cite{wang2018learning}                     & 33.8              & 35.9            & 99.5             & 98.4           & -                 & -               & -               & -               \\
LTCC\cite{qian2020long}                        & 34.38             & -               & 64.2             & -              & -                 & -               & -               & -               \\
FSAM\cite{hong2021fine}                    & 54.5              & -               & 98.8             & -              & 78.6              & 78.9            & 94.7            & 94.8            \\
SGPS\cite{shu2021semantic}                    & 65.8              & 61.2            & 99.5             & 96.7           & -                 & -               & -               & -               \\
Sync-Person-Cloud                    & 43.7              & 39.8            & 87.4             & 82.1           & 67.4              & 62.5            & 91.9            & 89.3            \\ \hline
\end{tabular}
\label{tab:SOA}
\end{table*}

\section{Experiments}
\subsection{Implementation Details}

We implement our model with Pytorch (\cite{paszke2019pytorch}). We conduct our model based on the unsupervised learning baseline \cite{dai2021cluster}. We adopt the ResNet-50 (\cite{he2016deep}) as our backbone. We modify the last layer stride to be 1 in the backbone Resnet50 to make final output features have more abundant information. Then we extract 2048d features for all images from the GAP layer in our backbone. During testing, we take the 2048d features to calculate the distance. For the beginning of each epoch, we use DBScan as our clustering method to generate pseudo labels for unlabeled input images.

The input image is resized to $256 \times 128$. The random cropping and horizontal flipping are performed as the augmentation methods. We haven't used random erasing and image padding because we want to keep the content of the input image to get synthetic clothing change person images. 
The batch size $P \times K$ is set as 32, in which $K$ is set to 4 as person images of identity and $P$ is set to 8 as pseudo person identities. The synthetic person image number for each input image $S$ is set to 4. So we process 160 images per mini-batch. We set the weight factor of loss $\alpha$ as 0.3. We set the momentum for update the memory instance $m$ as 0.3. 

We use Adam optimizer and set both the weight decay factor and weight decay bias factor as 0.0005. The base learning rate is 0.00035. We use a linear warming up strategy for adjustment of learning rate at train stage. For the first $E_{warmup}$ epoch with start learning rate $R_{start}$, the learning rate will linearly increase from $\frac{R_{start}}{E_{warmup}}$ to $R_{start}$. We set warm-up iteration number $E_{warmup}$ as 20 in our experiments. The total training epoch number is 120. 
We set the maximum distance $\varepsilon$ between two samples as 0.4 and the minimal number of neighbors in a core point $M$ as 4 for our clustering method DBScan hyperparameters.

\subsection{Datasets and Evaluation Protocol}

We mainly evaluated our method on two cloth changing Re-ID datasets: PRCC \cite{yang2019person} and VC-Clothes \cite{wan2020person}.

PRCC is named Person Re-id under the moderate Clothing Change dataset. It was collected for the task of cloth-changing person re-ID. PRCC consists of 221 identities captured by three camera views. VC-Clothes is a synthetic dataset rendered by the GTA5 game engine, which contains 19060 images of 512 identities captured from 4 cameras.

For evaluation protocol, we adopted the mean average precision (mAP) and ard Cumulative Match-ing Characteristics (CMC) rank-k accuracy.
For both cloth-changing datasets PRCC and VC-Clothes, we followed their evaluation protocols and evaluated the performance. For PRCC, we used single-shot matching by randomly choosing one image of each identity as the gallery. The cloth-changing setting in PRCC means there are all clothing change samples in the test set. For VC-Clothes, we used multi-shot matching and the clothing change setting is the same as that of PRCC. We also do the experiments on both datasets under the same clothing settings, whose test sets are all cloth-consistent samples in the test set.

\subsection{Comparison With the State-of-The-Art}

As there is no existing unsupervised clothing change person ReID method, we compare our method with the state-of-the-art supervised clothing change person ReID method.
As shown in Table \ref{tab:SOA},  we can see that our method is not the highest solution overall methods. The reason is that although we use person augmentation information and a strong unsupervised deep ReID pipeline, our model still lacks some supervised information to converge the ReID backbone. Through self identity constrain loss for synthetic clothing change person image feature and using a clustering algorithm to get pseudo labels, we can get some identity information from the dataset. However, the pseudo label is inaccurate and the person number keeps change so that we can't use classification loss. The synthetic clothing change image is generated from a pretrained $PISE$ module, whose image information is from other texture style transfer datasets. Hence, the quality of the supervised information restricts our model's ability. 

However, our method can perform as well as some supervised person ReID model, which is not design for clothing change scenarios. As shown in the table \ref{tab:SOA}, our method have higher Rank 1 accuracy and mAP than PCB \cite{sun2018beyond}, RGA-SC \cite{zhang2020relation} and LOMO \cite{liao2015person}. Hence, our method is effective for unsupervised clothing change ReID.

\subsection{Ablation Study}
\subsubsection{Component Effectiveness}
In this section, we study the effectiveness of key components of our proposed method. We take PRCC dataset under clothing change settings as an example. The baseline is a pure unsupervised person ReID pipeline. The components include $C_{a}$, $C_{i}$ and $C_{s}$. $C_{a}$ is the $PISE$ person clothing change image augmentation module, $C_{i}$ is the self identity constrain loss and $C_{s}$ is the cluster average and hardest sample operation. Because the $C_{i}$ must be based on $C_{a}$, we need to at least use both components when we want to analyze the effectiveness of $C_{i}$. 
The results of our proposed different components' effectiveness are shown in Table \ref{tab:ablation}.

\begin{table}[h]
\centering
\caption{Self comparison on PRCC dataset. }
\begin{tabular}{ccclcc}
\hline
\multirow{2}{*}{Method} & \multicolumn{3}{c}{Component}    & \multicolumn{2}{c}{PRCC} \\ \cline{2-6} 
                        & \multicolumn{1}{l}{$C_{a}$} & $C_{i}$ & $C_{s}$ & R1          & mAP        \\ \hline
Baseline                &                        &    &    &            29.6        & 27.4             \\
Baseline                & \checkmark                        &    &    &             30.3        & 27.9            \\
Baseline                &                        &    & \checkmark   &            32.4        & 30.8             \\
Baseline                & \checkmark                        &    & \checkmark   &            35.1        & 33.3            \\
Baseline                & \checkmark                        & \checkmark    &   &   37.2        & 35.3            \\
Sync-Person-Cloud                    & \checkmark                        & \checkmark    & \checkmark    & 43.7        & 39.8       \\ \hline
\end{tabular}
\label{tab:ablation}
\end{table}

As shown in Table \ref{tab:ablation}, our proposed model with all components significantly outperformed the baseline model. The baseline gets 29.6\% R1 accuracy, which is much lower than the most of state of the art method. By using $C_{a}$, we can get much more information from the augmented clothing change person images. Through training these images, the model can improve 0.7\% R1 accuracy, which is not so significant. The reason for the low improvement is that although we use the augmentation module, the identity information is still not be constrained and with more images information it is hard for the model to converge. By using $C_{s}$, we can get 2.8\% R1 accuracy improvement. The hardest and average sampling can help the model with better and faster convergence. The average sampling can help the model learn more global information in the dataset and the hardest sampling can help the model learn from an extreme instance. Hence, by combining $C_{a}$ and  $C_{s}$ together, we can get 35.1\% in R1 accuracy. Through using $C_{i}$ to constrain the synthetic person image features generated from the same original input person image, the model can get 37.2\% R1 accuracy. This significant improvement shows that the augmentation images have abundant person identity information for model disentanglement between people's clothing and identity, which helps our model to adapt the clothing change person ReID scenario. Finally, our method with all components can reach the highest R1 accuracy 43.7\%, which means that the $C_{a}$, $C_{i}$ and  $C_{s}$ can work well corporately.

\subsubsection{Sample Method}

In this section, we want to discuss the effectiveness of our sampling method. We take PRCC dataset under clothing change settings as an example.  

\begin{table}[h]
\centering
\caption{Sampling method comparison on PRCC dataset. }
\begin{tabular}{ccc}
\hline
\multirow{2}{*}{Sampling Method} & \multicolumn{2}{c}{PRCC}                    \\ \cline{2-3} 
                               & R1                   & mAP                  \\ \hline
No Sample                      & 37.2                 & 35.3                     \\
Average Sample                 & 36.1                 & 34.7                      \\
Hardest Sample                 & 41.2                 & 37.4                  \\
Both                           & 43.7                 & 39.8                 \\ \hline
\end{tabular}
\label{tab:sample}
\end{table}

The comparison results are shown in Table \ref{tab:sample}. All the settings are based on our model with image augmentation and self identity constrain. The first setting is that we don't use sampling in our contrast module, which means the query features and memory need to update one by one. In this case, the feature memory is not a single instance for each identity. It will record multiple instances in the memory to decrease the update frequency for each instance. It can achieve 37.2\% in Rank1 accuracy. Using the average sampling method decreases the accuracy by 1.1\% than first settings. 
The reason is mainly that although we use the average sample to reduce the memory load and utilize the global information of the same pseudo person, the model is harder to converge than use all the instances one by one. Another reason is that the pseudo label is not stable for update global information, which also causes low performance. By using the hardest sampling method, we can get 41.2\% Rank1 accuracy. The hardest sampling can help the model better converge and learn from the extreme image features. By using both average and hardest sampling, our model can get the highest Rank1 accuracy 43.7\%. In this situation, using global information is not the obstruction of the model convergence with the help from the hardest sampling. The hardest sampling will lead the gradient to the extreme instance and average sampling will help the model keep the global information of all images (including the synthetic clothing change images).
\subsubsection{DBScan Hyper-parameters}

In the DBScan clustering algorithm, $\varepsilon$ is the maximum distance between two samples for one to be considered as in the neighborhood of the other. It greatly affects the performance of clustering by deciding the final number of clusters (person pseudo identity numbers in our case).

When $\varepsilon$ is too small, the cluster with a small density will be divided into multiple clusters with similar properties. However, when $\varepsilon$ is too large, clusters with a closer distance and a larger density will be merged into one cluster. In the case of high-dimensional data in the person ReID problem, it is more difficult to select the $\varepsilon$ value due to the curse of dimensionality.  We analyze $\varepsilon$ influence on the PRCC dataset under clothing change settings using our whole model.

\begin{table}[h]
\centering
\caption{Analysis of $\varepsilon$ value on PRCC dataset.}
\begin{tabular}{ccc}
\hline
\multirow{2}{*}{$\varepsilon$ Value}  & \multicolumn{2}{c}{PRCC}                            \\ \cline{2-3} 
                        & R1                       & mAP                      \\ \hline
0.2                     & 38.4                     & 33.1                     \\
0.3                     & 41.2                     & 36.5                     \\
0.4                     & 43.7                     & 39.8                     \\
0.5                     & 40.8                     & 35.9                     \\
0.6                     & 40.2                     & 35.8                     \\
0.7                     & 40.4                     & 34.1                     \\ \hline
\end{tabular}
\label{tab:value}
\end{table}

As shown in the Table \ref{tab:value}, we can see that our method works best when $\varepsilon$ equals 0.4.

\section{Conclusion}
In this paper, to solve the lack of person identity label and clothing change problem in person ReID, we propose an unsupervised clothing change person ReID model called Sync-Person-Cloud. To our best knowledge, we are the first to use unsupervised learning in the clothing change ReID problem. To improve the ability of extraction clothing change invariant information of our model, we propose a self identity constrain for synthetic clothing change person image and a hardest and average sampling method for cluster contrast. 
Experiments show the performance of our model compared with the state-of-the-art supervised clothing change person ReID methods.

\bibliography{aaai22}

\end{document}